\documentclass[runningheads]{llncs}
\usepackage[T1]{fontenc}
\usepackage{graphicx}
\usepackage{listings}

\begin{document}
%
\title{Bridging Static and Agentic RAG for Taiwanese Historical Question Answering}
\titlerunning{Bridging Static and Agentic RAG for Taiwanese Historical QA}
%
%

\author{
Kai-Hsin Chen\inst{1}\thanks{Equal contribution with Wei-Yu Chen; corresponding author.}
\and
Wei-Yu Chen\inst{2}
\and
Xuanjun Chen\inst{3}
\\
Su-Ling Yeh\inst{4}
\and
\mbox{Jyh-Shing Roger Jang}\inst{1,2}
}

\authorrunning{K.-H. Chen et al.}

\institute{
Graduate Institute of Networking and Multimedia,
National Taiwan University
\and
Dept. of Computer Science \& Information Engineering,
National Taiwan University
\and
Graduate Institute of Communication Engineering,
National Taiwan University
\and
Dept. of Psychology, National Taiwan University
\\
\email{r14944026@csie.ntu.edu.tw}
}

%
\maketitle              
\begin{abstract}

Agentic retrieval-augmented generation (RAG) enables language models to adapt retrieval based on previously retrieved evidence, but it remains unclear whether such adaptive orchestration consistently outperforms well-designed static pipelines. We conduct a controlled comparison of agentic and static RAG for Taiwanese historical question answering, sharing the same generator and hybrid retrieval backend. Despite similar aggregate performance, the two pipelines differ on 70.83\% of questions, with their advantages largely canceling out when averaged. An oracle that selects the better response per question improves the composite score by 0.2417 over the better individual pipeline, revealing substantial headroom for question-level selection. We therefore introduce a post-hoc selector that compares the two responses and their cited evidence, significantly outperforming either individual pipeline and recovering 60.34\% of the oracle headroom. These results show that aggregate comparisons can obscure meaningful
question-level differences between retrieval strategies, suggesting that
exploiting their complementarity may be more fruitful than seeking a
universally superior pipeline.

\keywords{Retrieval-Augmented Generation \and Agentic RAG \and
Retrieval Orchestration \and Historical Question Answering \and
Answer Selection}
\end{abstract}
\section{Introduction}

Knowledge-intensive question answering often relies on external evidence,
making the orchestration of retrieval a central design choice in
retrieval-augmented generation (RAG) systems. Static RAG follows a
predefined retrieval procedure, whereas agentic RAG allows an LLM agent
to adapt its queries and invoke additional retrieval based on the evidence
obtained during generation. Although such flexibility may better
accommodate the needs of individual questions, it remains unclear whether
agent-controlled retrieval consistently improves answer quality over a
well-designed static pipeline. In Taiwanese historical question answering, prior work has
demonstrated the effectiveness of RAG using predefined retrieval
pipelines~\cite{lin2025taiwanrag}, but whether adaptive,
agent-controlled retrieval provides additional benefits over
predefined retrieval remains unexplored.

To investigate this question, we compare agentic and static RAG while
sharing the same generator and hybrid retrieval backend, allowing us to
focus the comparison on retrieval orchestration. Beyond aggregate
performance, we examine whether the two pipelines exhibit complementary
advantages across individual questions and whether these differences can
be exploited through per-question selection. We further introduce a
post-hoc selector that compares the two responses and their cited evidence
to identify the preferred response.

Our results yield three main findings. First, agentic RAG does not
provide a consistent aggregate advantage over static RAG. Second,
this aggregate similarity masks substantial question-level differences
between the two retrieval strategies, indicating that their relative
strengths vary across questions. Third, our post-hoc selector exploits
these differences and significantly outperforms either individual
pipeline. Together, these findings suggest that, rather than seeking
a universally superior retrieval strategy, a more promising direction
is to determine which strategy is better suited to each question.
This perspective further motivates routing between
retrieval orchestration strategies before generation. 

\section{Related Work}



\subsection{Retrieval-Augmented Generation}

Retrieval-augmented generation (RAG) grounds language models in external knowledge for complex NLP tasks~\cite{lewis2020rag}. Standard methods often treat retrieved documents as isolated units, which limits multi-hop reasoning. To overcome this, recent frameworks like CodaRAG~\cite{li2026codarag} reconstruct dispersed evidence chains via active associative discovery. In the domain of Taiwanese historical QA, Lin et al.~\cite{lin2025taiwanrag} previously explored how query traits and metadata impact RAG performance. Building on this setting, we investigate retrieval orchestration, comparing agent-controlled versus predefined retrieval under a shared backbone. 

\subsection{Agentic and Adaptive Retrieval}

Prior work has explored iterative interaction between language models
and external information sources. 
Iterative approaches like ReAct~\cite{yao2023react} and IRCoT \cite{trivedi2023ircot} interleave reasoning and retrieval to build upon previous observations. To mitigate error propagation and over-generation in these multi-step systems, GDP-RAG~\cite{chou2026only} plans efficient trajectories by targeting only the missing information delta. Furthermore, recent methods dynamically adapt retrieval behavior: Adaptive-RAG \cite{jeong2024adaptive} selects strategies based on question complexity, while Ferrazzi et al.~\cite{ferrazzi2026agentic} employ LLMs to autonomously orchestrate retrieval decisions and stopping criteria. 

Our work complements these studies by focusing on a controlled
comparison of agent-controlled and predefined retrieval under a shared
retrieval and generation backbone in Taiwanese historical QA. We further
examine whether their question-level differences can be exploited through
post-hoc answer selection.

\subsection{RAG Evaluation and Answer Selection}

Recent RAG evaluation frameworks provide fine-grained diagnostics
beyond aggregate end-to-end scores. RAGChecker~\cite{ru2024ragchecker},
for example, evaluates retrieval and generation through claim-level
metrics, helping expose behaviors that aggregate scores may obscure.
Our analysis considers a complementary axis of variation: how the
relative performance of different retrieval strategies changes across
individual questions, even when their aggregate performance is similar.

Related work has explored both post-generation response selection and
pre-generation routing. LLM-Blender~\cite{jiang2023llmblender} ranks
candidate responses from multiple LLMs after generation, whereas
RAGRouter~\cite{zhang2025ragrouter} routes queries among different RAG
models before generation. Our setting differs in that the candidate
responses arise from different retrieval orchestration strategies while
sharing the same generator and retrieval backend. We quantify the
selection headroom arising from these question-level differences and
evaluate how much of it can be recovered by a post-hoc selector using
the responses and their cited evidence.

\section{Methods}

\begin{figure}[t]
\centering
\includegraphics[width=\textwidth]{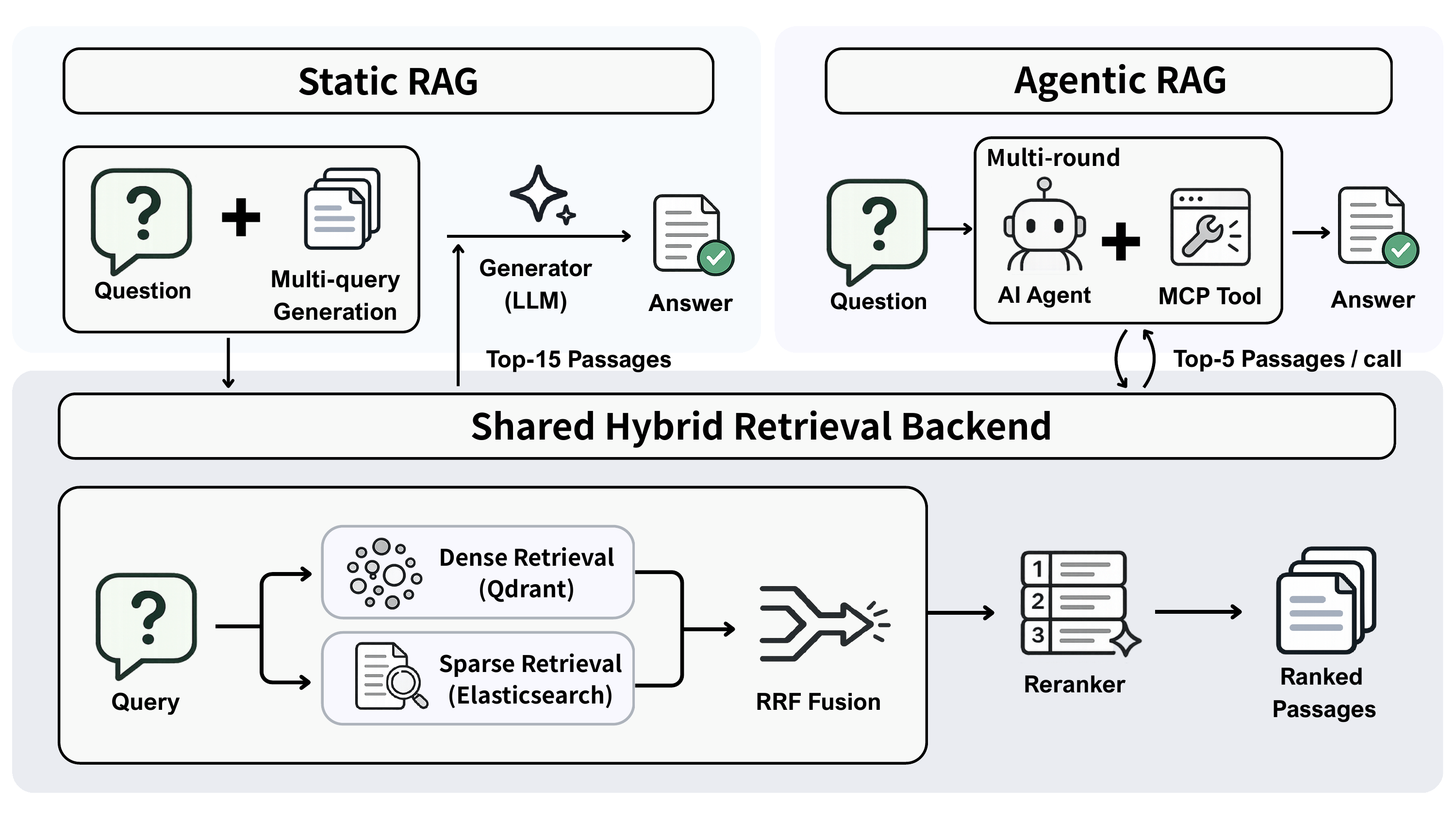}
\caption{Comparison of Static and Agentic RAG under a shared
hybrid retrieval backend. Static RAG follows a predefined
multi-query retrieval procedure and provides the top 15 passages to
the generator, whereas Agentic RAG adaptively invokes the same
retrieval backend through an MCP tool and receives the top five
passages per call.}
\label{fig:method_overview}
\end{figure}

We compare Static and Agentic RAG under a shared retrieval and generation backbone, isolating differences in retrieval orchestration (Fig.~\ref{fig:method_overview}). We further investigate whether their question-level performance differences can be exploited through post-hoc answer selection (Fig.~\ref{fig:selector_framework}).

\subsection{Shared Retrieval and Generation Backbone}

Both RAG pipelines share the same retrieval backend and answer
generator. Given a search query, the hybrid backend retrieves candidate
passages using dense and sparse retrievers. Their ranked lists are
combined using Reciprocal Rank Fusion (RRF)~\cite{cormack2009reciprocal},
followed by a cross-encoder reranker that produces the final ranking.
The pipelines differ primarily in how and when this shared retrieval
backend is invoked, as described next.

\subsection{Retrieval Orchestration}

In Static RAG, the question is first transformed into multiple search
queries, which are submitted to the shared retrieval backend. The top-15 reranked passages are then provided to the generator as context for
the final answer. The retrieval procedure is therefore fixed before
answer generation, with no mechanism to revise subsequent retrieval
based on the retrieved evidence.

In Agentic RAG, retrieval is instead controlled by the LLM agent. The
agent accesses the same retrieval backend through an MCP retrieval tool
and determines both the search queries and whether additional retrieval
is needed. Each retrieval call returns the top-5 reranked passages,
which remain available in the agent's context for subsequent reasoning
and retrieval decisions. The agent may therefore perform multiple
retrieval calls before producing its final answer, allowing subsequent
retrieval decisions to adapt to previously retrieved evidence.

\subsection{Post-hoc Answer Selection}

\begin{figure}[t]
\centering
\includegraphics[width=\textwidth]
{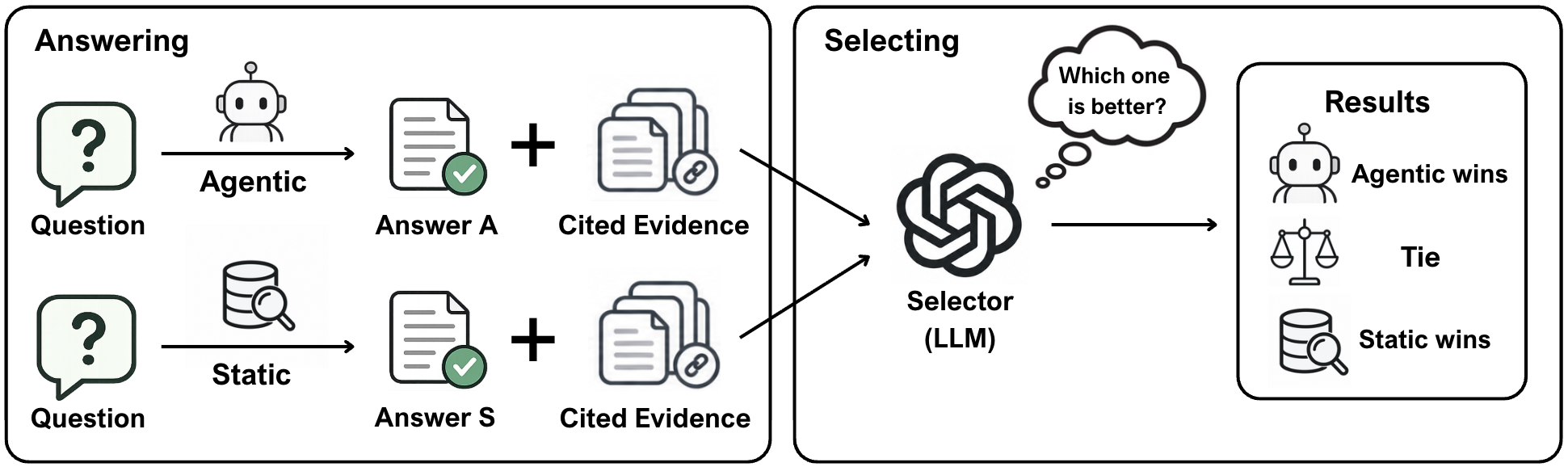}
\vspace{-0.5em}
\caption{Post-hoc answer selection framework. Agentic and Static RAG independently generate answers with cited evidence, which are compared by an LLM-based selector to determine the preferred answer or a tie.}
\label{fig:selector_framework}
\end{figure}

To test whether these question-level differences can be exploited, we
evaluate a post-hoc selector that chooses between the independently
generated responses of the two RAG pipelines. As illustrated in
Fig.~\ref{fig:selector_framework}, the selector receives the original question,
the two candidate responses, and the union of their cited evidence.
The comparison considers each response's support from the provided
evidence, as well as its relevance and completeness with respect to
the question.

To reduce sensitivity to response ordering, each pair is evaluated twice
with the candidate positions reversed. A response is selected only when
the two evaluations agree on the same candidate; otherwise, the selector
returns a tie (i.e., abstains from choosing either response). Tie cases
remain included in the evaluation, with their scoring treatment described
in Section~\ref{sec:evaluation}. The complete selector prompt is provided in Appendix~\ref{app:selector_prompt}.

\section{Experimental Setup}

We evaluate the two retrieval orchestration paradigms primarily on
open-ended historical question answering, enabling multidimensional
analysis of generation quality. We additionally use an objectively
scored QA benchmark to complement this analysis with a direct measure
of answer correctness.

\subsection{Datasets and Benchmarks}

\paragraph{Ming-Qing Taiwan Government Administrative Archives.}
Our primary evaluation uses the Ming-Qing Taiwan Government
Administrative Archives (Minqing), which contains 46,709 historical
documents. The benchmark consists of 120 open-ended questions derived
from these archives, with each question associated with a single gold
document used as its reference evidence.

\paragraph{Taiwanese Folklore and Cultural Artifacts Dictionary.}
As a supplementary objective evaluation, we use 102 questions from
the Taiwanese Folklore and Cultural Artifacts Dictionary (TFCAD),
consisting of 52 fill-in-the-blank and 50 multiple-choice questions.
Unlike the open-ended Minqing benchmark, these questions permit
objective correctness evaluation and provide a complementary measure
that does not rely on generation-quality judgments.

\paragraph{Corpus preprocessing.}
Both corpora are processed using the same chunking and indexing pipeline.
Documents are segmented at sentence boundaries and greedily packed into
non-overlapping chunks of at most 500 characters, which we treat as
retrieval passages. This results in 117,786 passages from 46,709 Minqing
documents and 3,121 passages from 2,805 TFCAD entries. Retrieval and
reranking operate at the passage level, while retrieved evidence is
assembled at the document level for downstream generation.

\subsection{Implementation Details}

The shared hybrid retrieval backend uses
Qwen3-Embedding-0.6B~\cite{qwen3embedding} with
Qdrant~\cite{qdrant} for dense retrieval and
BM25~\cite{robertson2009probabilistic} with
Elasticsearch~\cite{elasticsearch} for sparse retrieval. The dense and sparse rankings are combined using
RRF~\cite{cormack2009reciprocal} with $k=60$ and reranked using
BGE-reranker-large~\cite{xiao2023cpack}. Static RAG provides the
top-15 reranked passages to the generator, whereas each retrieval call
in Agentic RAG returns the top-5 reranked passages. Both pipelines use
Claude Sonnet 5~\cite{anthropic2026sonnet5} as the answer generator. 
Agentic RAG accesses the retrieval backend through the
\texttt{retrieve\_documents} MCP tool and may invoke it multiple times
before producing the final answer. The post-hoc selector uses GPT-5.5~\cite{openai2026gpt55} for pairwise answer
comparison. For each question, the selector receives all passages cited
by either candidate response; the largest evidence set contains 36
passages.

\subsection{Evaluation}
\label{sec:evaluation}
\paragraph{Open-ended generation quality.}
Drawing on prior multidimensional RAG evaluation
frameworks~\cite{ru2024ragchecker,es2024ragas}, we evaluate each
Minqing response along four dimensions: Correctness, Completeness,
Faithfulness, and Relevance. Because the benchmark provides gold
documents but no human-written reference answers, we generate a
reference answer for each question using Claude Opus 5~\cite{anthropic2026opus5},
conditioned on the question and its gold document. Claude Opus
4.8~\cite{anthropic2026opus48} then evaluates each response using the reference answer
and the evidence cited by the corresponding RAG pipeline. Each dimension is scored from 1 to 5, and their mean defines the
composite generation-quality score.

\paragraph{Selector evaluation.}
For each question, selector performance is measured by the composite
generation-quality score of the selected response. When the selector
returns a tie, we assign the mean of the two candidate scores,
corresponding to their expected value under uniform selection. We compare
the selector against random selection, which chooses between the two
responses with equal probability, and a length heuristic, which selects
the longer response. We additionally report an oracle that selects the
higher-scoring response for each question, defining the available
selection headroom. We quantify how much of this headroom is captured
by a selection strategy as

\begin{equation}
\mathrm{Recovery} =
\frac{S_{\mathrm{sel}} - S_{\mathrm{best}}}
{S_{\mathrm{oracle}} - S_{\mathrm{best}}}
\times 100\%.
\label{eq:recovery}
\end{equation}

where $S_{\mathrm{sel}}$ is the score achieved by a selection strategy,
$S_{\mathrm{best}}$ is the score of the better individual pipeline, and
$S_{\mathrm{oracle}}$ is the oracle score. Thus, the better individual
pipeline corresponds to 0\% recovery and the oracle to 100\%. We further
examine sensitivity to alternative tie-handling strategies in Section~\ref{sec:tie_sensitivity}.

\paragraph{Statistical analysis.}
To account for question-level variability, we report 95\% confidence
intervals for paired score differences using paired bootstrap resampling
with 20,000 resamples~\cite{efron1993bootstrap}. We use the Wilcoxon
signed-rank test~\cite{wilcoxon1945individual} to test paired differences
in generation-quality scores. For TFCAD, we use the exact McNemar
test~\cite{mcnemar1947sampling} for paired binary correctness.

\section{Results}
We first compare the overall generation quality of Agentic and Static RAG, then examine question-level differences and post-hoc selection performance. Finally, we analyze retrieval behaviors associated with these differences.

\subsection{Overall Generation Quality}
\label{sec:overall}

Table~\ref{tab:main_results} compares the two retrieval orchestration
strategies on the Minqing benchmark. Agentic RAG achieves a slightly
higher composite score than Static RAG (4.2104 vs.\ 4.1188), but this
numerical advantage is not statistically significant
($\Delta=+0.0917$, 95\% CI: $[-0.0854,\,0.2687]$; $p=0.6543$).
Thus, Agentic RAG does not provide a reliable overall improvement over
the Static retrieval pipeline when evaluated by aggregate generation
quality.

The dimensional results further show that the absence of an overall
advantage does not imply identical behavior across evaluation criteria.
Agentic RAG is numerically higher in correctness, completeness, and
relevance, but only the relevance improvement is statistically
significant ($\Delta=+0.1750$, 95\% CI:
$[+0.0250,\,+0.3250]$; $p=0.0331$). In contrast, Static RAG is
slightly higher in faithfulness, although the difference is not
significant. These results suggest that the observed advantage of
Agentic RAG is dimension-specific rather than a general improvement
in answer quality. This aggregate pattern is also consistent with the
supplementary judge-free evaluation on TFCAD, where Agentic RAG is
numerically higher but the difference is not significant
(Appendix~\ref{app:tfcad}).

The aggregate results, however, leave open two distinct possibilities.
The two pipelines may perform similarly on most questions, or they may
excel on different questions such that their advantages cancel out when
averaged. Aggregate scores alone cannot distinguish between these
explanations. We therefore next examine their paired performance at the
individual-question level.

\begin{table}[t]
\centering
\caption{Generation quality of Agentic and Static RAG on the
Minqing benchmark ($n=120$). $\Delta$ denotes Agentic minus Static;
$p$-values are from two-sided Wilcoxon signed-rank tests.}
\vspace{-0.5em}
\label{tab:main_results}

\small
\setlength{\tabcolsep}{2.2pt}
\renewcommand{\arraystretch}{1.15}

\begin{tabular}{@{}lccccc@{}}
\noalign{\hrule height 0.9pt}

& \textbf{Correctness}
& \textbf{Completeness}
& \textbf{Faithfulness}
& \textbf{Relevance}
& \textbf{Composite} \\

\noalign{\hrule height 0.4pt}

Static RAG
& 4.1750
& 3.9833
& 4.5417
& 3.7750
& 4.1188 \\

Agentic RAG
& 4.2917
& 4.1167
& 4.4833
& 3.9500
& \textbf{4.2104} \\

\noalign{\vskip 2pt}
\hline
\noalign{\vskip 2pt}

$\Delta$ (A$-$S)
& +0.1167
& +0.1333
& $-0.0583$
& \textbf{+0.1750}
& +0.0917 \\

Wilcoxon $p$
& 0.2927
& 0.4204
& 0.4304
& \textbf{0.0331}
& 0.6543 \\

\noalign{\hrule height 0.9pt}
\end{tabular}

\end{table}

\subsection{Question-level Performance Differences}

\begin{figure}[t]
    \centering
    \includegraphics[width=\textwidth]
    {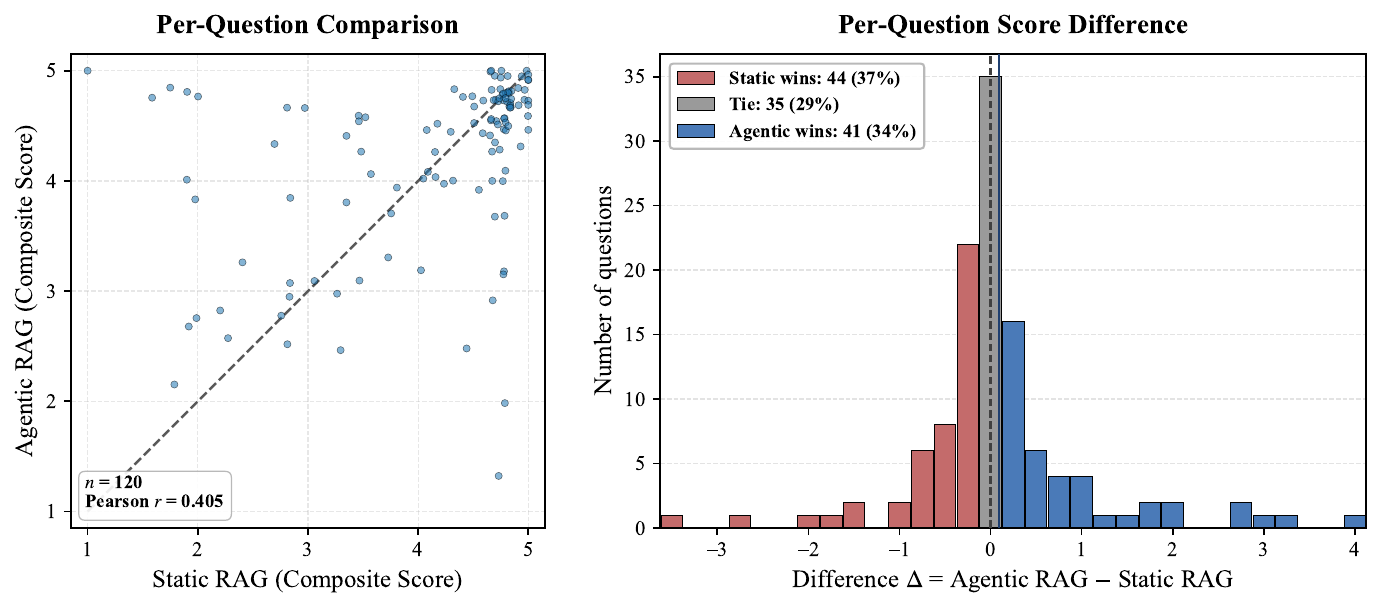}
    \vspace{-0.5em}
    \caption{Question-level comparison of Agentic and Static RAG on
    Minqing: paired composite scores (left) and Agentic-minus-Static score
    differences (right).}
    \label{fig:question_level_comparison}
\end{figure}

The aggregate similarity between the two pipelines masks substantial
differences at the individual-question level. As shown in
Fig.~\ref{fig:question_level_comparison}, the paired composite scores
are distributed on both sides of the diagonal, and the distribution
of score differences similarly shows substantial variation in both
directions. Agentic RAG achieves a higher score on 41 questions,
Static RAG on 44, and the two obtain identical scores on 35. Overall,
the pipelines differ on 85 of 120 questions (70.83\%), while their
wins are nearly evenly split. The aggregate similarity therefore
reflects opposing question-level advantages that largely cancel out
when averaged, rather than uniformly similar performance across
questions.

These complementary question-level advantages create substantial
potential for per-question selection. An oracle that selects the
higher-scoring response for each question achieves a composite score
of 4.4521, compared with 4.2104 for the better individual pipeline,
yielding a headroom of $+0.2417$. This headroom is 2.64 times larger
than the aggregate Agentic--Static difference, motivating a practical
question: how much can a selector recover using only the candidate
responses and their cited evidence?

\subsection{Post-hoc Answer Selection}

Table~\ref{tab:selector_results} reports the performance of the
post-hoc selector and comparison strategies. The selector achieves
a composite score of 4.3563, improving over the better individual
pipeline, Agentic RAG, by $+0.1458$. This improvement is statistically significant
(95\% CI: $[+0.0604,\,+0.2469]$; $p=0.0012$) and recovers
60.34\% of the available headroom toward oracle performance.

\begin{figure}[t]
    \centering
    \includegraphics[width=0.88\textwidth]
    {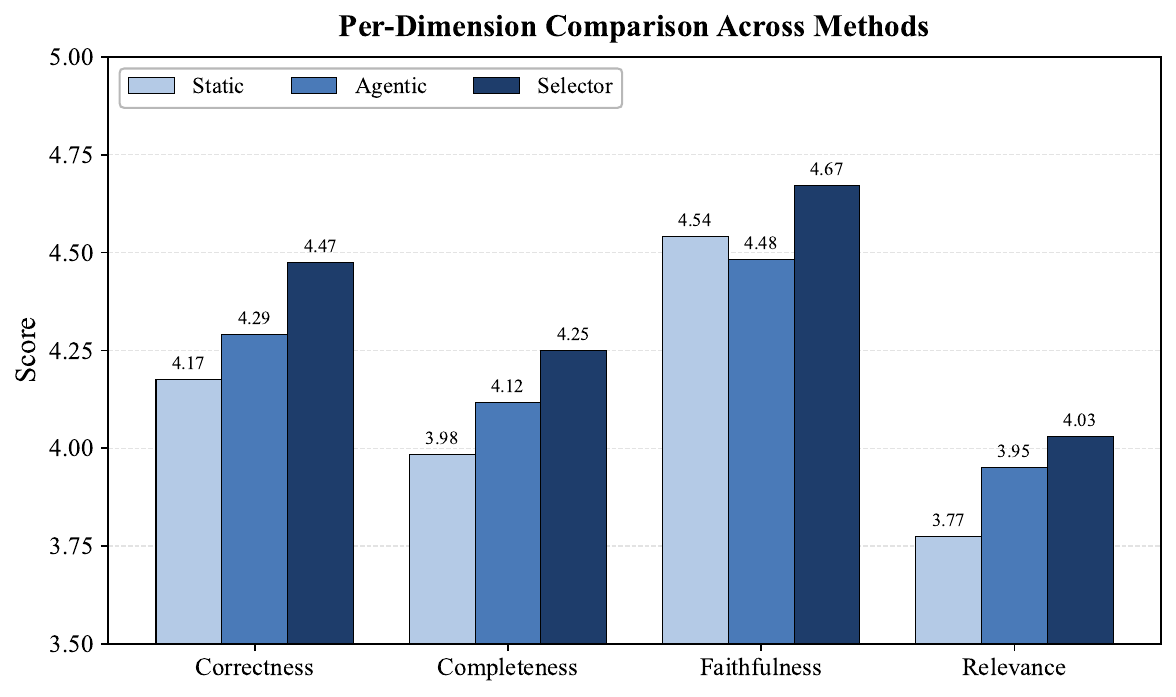}
    \caption{Generation quality across evaluation dimensions for Static
    RAG, Agentic RAG, and the post-hoc selector on the Minqing benchmark.}
    \label{fig:selector_dimensions}
\end{figure}

The selector achieves the highest mean score across all four
dimensions (Fig.~\ref{fig:selector_dimensions}) and outperforms both
random selection and the length heuristic
(Table~\ref{tab:selector_results}). Notably, the length heuristic
identifies the better response on only 47.06\% of questions with
unequal pipeline scores. These results indicate that simple rules
based on random choice or response length cannot account for the
selector's gains.

Among questions with unequal pipeline scores, the selector makes a
decisive choice in 67 cases and selects the better response in 50 of
them (74.63\%). Correct selections yield an average gain of 1.07
points, whereas incorrect selections incur an average loss of 0.43
points. The benefit of a correct decision is, therefore, approximately
2.5 times the cost of an incorrect one. This asymmetry helps explain
why imperfect selection can still produce a substantial aggregate
improvement: correct decisions tend to capture larger quality
differences than those lost through incorrect decisions on other questions.

\begin{table}[t]
\centering
\caption{Performance of post-hoc selection strategies on the Minqing
benchmark. Recovery denotes the percentage of oracle headroom recovered
relative to the better individual pipeline; negative values indicate
performance below this baseline.}
\vspace{-0.5em}
\label{tab:selector_results}
\begin{tabular}{lcc}
\noalign{\hrule height 0.8pt}
\textbf{Strategy} & \textbf{Composite} & \textbf{Recovery (\%)} \\
\hline
Static RAG & 4.1188 & -37.93 \\
Agentic RAG        & 4.2104 & 0.00 \\
Random selection   & 4.1646 & -18.97 \\
Length heuristic   & 4.1917 & -7.76 \\
\textbf{Post-hoc selector} & \textbf{4.3563} & \textbf{60.34} \\
Oracle             & 4.4521 & 100.00 \\
\noalign{\hrule height 0.8pt}
\end{tabular}
\end{table}

\subsection{Analysis of Agentic Retrieval Behavior}

To better understand when adaptive retrieval is associated with
improved performance, we examine the retrieval behavior of Agentic
RAG. The agent performs multiple retrieval calls on 68 of 120
questions (56.67\%), with each multi-call case involving query
reformulation rather than simple retries. Among questions with unequal
pipeline scores, Agentic wins exhibit greater query diversity than
Agentic losses, as measured by token-level Jaccard overlap between
successive queries (0.076 vs.\ 0.141). In contrast, the contribution
of later-round evidence is nearly identical between the two groups.
Together, these results suggest that successful iteration is associated
more with redirecting the search than with simply retrieving additional
evidence.

This behavior is particularly relevant when the initial retrieval
misses the required evidence. On 18 of 120 questions (15.0\%), the
gold document enters the Agentic context only after the first retrieval
round; on these questions, Agentic RAG achieves a mean composite
advantage of $+0.54$ over Static RAG. This result provides a concrete case in which iterative retrieval can
recover evidence missed during the initial retrieval round. 

\paragraph{Case studies.}
Two contrasting cases further illustrate how adaptive retrieval can
produce large performance differences in either direction. In Q97
(Agentic 5.00, Static 1.00), the question asks about an edict relayed
``on the ninth day of the fifth month.'' The relevant memorial is
split across two chunks: one contains the date and the requested edict,
whereas the other contains only the submission date of the twelfth.
Because the Static pipeline deduplicates retrieved chunks by document
identifier during hybrid fusion and multi-query merging, only one chunk
from the memorial can remain in its context. In this case, it retained
the second chunk, which contains none of the requested edict, and
therefore concluded that no such record existed. Agentic RAG initially
reached the same uninformative chunk in its third retrieval call, but
continued searching: it reformulated the query with the target date
restored and retrieved the relevant first chunk in its fourth call.
The additional retrieval round therefore corrected an initial
evidence-selection failure that the Static pipeline could not revisit.

Q29 (Agentic 2.00, Static 4.75) illustrates the corresponding failure
mode. The question asks for the net amount of customs revenue actually
transferred to Fujian; the correct amount, 340,000 taels after two
deductions, appears only in the gold memorial. During its second
retrieval call, the agent shifted the information need from the
``final actual total'' to the broader ``total amount allocated,''
leading it to retrieve a later and separate transfer. Because the
individual figures were supported by the retrieved evidence, the agent
stopped after two calls and incorrectly combined three transfers into
1,100,000 taels. In contrast, the Static pipeline generated multiple
query variants in a single round and reranked their pooled results,
placing both the gold memorial and a corroborating memorial in context.
It consequently reported 340,000 taels and correctly treated the later
transfer as a separate case. Together, these cases show that the flexibility enabling Agentic RAG
to recover from retrieval misses can also redirect the search toward
plausible but incorrect evidence. Its advantage and failure mode
therefore stem from the same adaptive mechanism: the agent controls
both how retrieval evolves and when the accumulated evidence is
sufficient to answer. This duality helps explain why greater retrieval
flexibility produces substantial question-level differences without
yielding a uniform aggregate advantage.

\section{Ablation and Robustness Analysis}

\subsection{Retrieval Ablation}
\label{sec:retrieval_ablation}

To contextualize the contribution of retrieval relative to retrieval
orchestration, we compare both RAG pipelines with a closed-book
baseline using the same generator (Claude Sonnet 5). On Minqing,
closed-book generation achieves a composite score of 2.3146, yielding
retrieval gains of $+1.8958$ for Agentic RAG and $+1.8042$ for
Static RAG; both improvements are statistically significant
($p<0.001$). The larger retrieval gain is 20.68 times the aggregate
Agentic--Static difference reported in Section~\ref{sec:overall}.
These results indicate that access to external evidence is associated
with substantially larger quality gains than the choice between
retrieval orchestration strategies in our setting.

\subsection{Sensitivity to Tie Handling}
\label{sec:tie_sensitivity}
The selector returns a tie for 25 of the 120 questions when the two
orderings do not yield a consistent preference. In the main
evaluation, these tie decisions are assigned the mean score of the
Agentic and Static responses, corresponding to their expected
value under uniform selection. We examine whether the selector's
improvement depends on this treatment using alternative tie-handling
rules together with upper and lower sensitivity bounds.

As shown in Table~\ref{tab:abstention_sensitivity}, resolving all ties
to either Agentic or Static RAG yields 56.90--63.79\% oracle
headroom recovery, and both settings remain significantly better than
Agentic RAG ($p\leq0.0015$). The difference between these two
tie-handling rules is itself not significant ($p=0.4252$), providing no evidence that either tie-handling rule is systematically preferable.

The sensitivity bounds yield the same conclusion. Even when every tie
is treated as incorrect, the selector remains above Agentic RAG
($+0.1104$; $p=0.0262$), while treating all ties as correct raises
oracle-headroom recovery to 75.00\%. Tie handling changes the
estimated magnitude of the gain, but does not determine whether the
selector outperforms the best individual pipeline. The main selection
result is thus robust to tie handling.

\begin{table}[t]
\centering
\caption{Sensitivity of selector performance to tie handling.}
\vspace{-0.5em}
\label{tab:abstention_sensitivity}
\begin{tabular}{lccc}
\noalign{\hrule height 0.8pt}
\textbf{Tie Handling}
& \textbf{Final Score}
& \textbf{Recovery (\%)}
& \textbf{vs.\ Agentic} \\
\hline

\multicolumn{4}{l}{\textit{Practical handling rules}} \\

Tie $\rightarrow$ Agentic
& 4.3479 & 56.90 & +0.1375 \\

Tie $\rightarrow$ Static
& 4.3646 & 63.79 & +0.1542 \\

\textbf{Expected Value (Main)}
& \textbf{4.3563} & \textbf{60.34} & \textbf{+0.1458} \\

\hline
\multicolumn{4}{l}{\textit{Sensitivity bounds}} \\

All Ties Correct
& 4.3917 & 75.00 & +0.1813 \\

All Ties Incorrect
& 4.3208 & 45.69 & +0.1104 \\

\noalign{\hrule height 0.8pt}
\end{tabular}
\end{table}

\section{Discussion}

Our results suggest that greater retrieval flexibility does not
translate into a uniform advantage in generation quality. Agentic RAG
can reformulate queries and recover evidence missed during the initial
retrieval round, providing a distinctive advantage on questions that
benefit from redirecting the search. However, these benefits are
conditional rather than universal: Static RAG remains competitive
overall and outperforms Agentic RAG on many individual questions.
Adaptive retrieval should therefore be viewed as a complementary
capability rather than a uniformly superior replacement for a
well-designed predefined retrieval pipeline.

The substantial question-level complementarity between the two pipelines motivates selecting retrieval strategies for individual questions rather than seeking a globally superior approach. Although our post-hoc selector exploits these differences, it requires both pipelines to generate responses. Pre-generation routing could potentially preserve these benefits more efficiently by selecting between adaptive and predefined retrieval before generation. Our findings motivate this direction but do not establish such a routing mechanism.

\section{Conclusion}

This work compared Agentic and Static RAG under a shared retrieval
and generation backbone for Taiwanese historical question answering.
We find that agent-controlled retrieval does not provide a consistent
aggregate advantage over a well-designed Static pipeline, despite
substantial differences at the individual-question level. These
differences create considerable headroom for per-question selection.
Our post-hoc selector exploits this complementarity, recovering
60.34\% of the available oracle headroom and significantly
outperforming either individual pipeline. Together, these results
highlight the importance of examining retrieval orchestration beyond
aggregate performance alone.

\paragraph{Limitations and future work.}
Several limitations should be considered when interpreting these
results. First, our open-ended evaluation relies on an LLM judge and
reference answers generated from gold documents, which may introduce
evaluation bias. Although the supplementary judge-free evaluation
provides supporting evidence, validation against human judgments would
further strengthen these findings. Second, the experiments are limited
to two Taiwanese historical QA benchmarks and a single generator
configuration. Although separate models are used for generation,
evaluation, and selection, the findings may still depend on the
particular model configuration and historical QA setting considered
here. Finally, the post-hoc selector requires both pipelines to generate responses before selection, increasing inference cost. Future work could explore pre-generation routing to select retrieval strategies more efficiently, potentially extending to other knowledge-intensive tasks, such as multi-hop question answering and scientific literature synthesis. Whether such routing generalizes beyond historical QA remains to be empirically validated.

\begin{credits}
\subsubsection{\ackname}
We gratefully acknowledge the support of the
National Science and Technology Council (NSTC),
Taiwan, under Grant No. NSTC 113-2740-H-002-001-MY3,
``TAIHUCAIS: TAIwan HUmanities Conversational AI
Knowledge Discovery System.''
This support has been instrumental in advancing
our research on retrieval-augmented generation
for Taiwanese historical question answering.
We also extend our sincere thanks to the members of the TAIHU benchmark team for their invaluable contributions: from National Taiwan University, Wei-Cheng Li, Pei-Hsuan Chang, and Hsing-Ju Lo; from National Central University, Wen-Jie Lu, Pin-Yi Li, and Richard Tzong-Han Tsai; from National Tsing Hua University, Ye-Cao Chen; and from Academia Sinica, Ting-Hsuan Lin.
\end{credits}

\bibliographystyle{splncs04}
\bibliography{references}
\lstdefinestyle{prompt}{
    basicstyle=\ttfamily\scriptsize,
    breaklines=true,
    breakatwhitespace=false,
    columns=fullflexible,
    keepspaces=true,
    showstringspaces=false,
    frame=single,
    aboveskip=4pt,
    belowskip=4pt,
    xleftmargin=2pt,
    xrightmargin=2pt
}
\appendix
\section{Supplementary TFCAD Results}
\label{app:tfcad}

\begin{table}[h]
\centering
\caption{Objective QA accuracy on TFCAD.}
\vspace{-0.5em}
\label{tab:tfcad_results}
\begin{tabular}{lccc}
\noalign{\hrule height 0.8pt}
\textbf{Question Type} & \textbf{Agentic} & \textbf{Static} & \textbf{$p$} \\
\hline
Fill-in-the-blank ($n=52$) & 0.9808 & 0.9615 & 1.0000 \\
Multiple-choice ($n=50$)   & 0.9400 & 0.8800 & 0.4531 \\
Overall ($n=102$)          & 0.9608 & 0.9216 & 0.3438 \\
\noalign{\hrule height 0.8pt}
\end{tabular}
\end{table}


Table~\ref{tab:tfcad_results} presents the supplementary TFCAD results using the exact McNemar test~\cite{mcnemar1947sampling} for paired correctness comparisons. Agentic RAG achieves higher accuracy than Static RAG, but the difference is not statistically significant overall (p = 0.3438) or within either question type. This finding is consistent with the Minqing results, where Agentic RAG shows no significant aggregate advantage.

\section{Evaluation and Selection Prompts}
\label{app:prompts}

\subsection{Generation-Quality Judge Prompt}
\label{app:judge_prompt}

The generation-quality judge receives the question, reference answer, retrieved context, and candidate answer. Its core instructions are summarized below, omitting implementation-specific details.

\begin{lstlisting}[
style=prompt,
caption={Core generation-quality evaluation instructions.},
label={lst:judge_prompt}
]
Evaluate the Candidate Answer using the Question,
Reference Answer, and Retrieved Context. Score each
dimension independently from 1 (poor) to 5 (excellent).

Correctness: Accuracy of substantive claims with
respect to the Reference Answer.
Completeness: Coverage of key information required
by the Reference Answer.
Faithfulness: Support for substantive claims in the
Retrieved Context. Do not use the Reference Answer
for this dimension.
Relevance: How directly the Candidate Answer
addresses the Question.

Do not use external knowledge or favor answers for
being longer or more detailed.
\end{lstlisting}






\subsection{Post-hoc Selector Prompt}
\label{app:selector_prompt}

The reference-free selector (Section~3.3) uses the
following system prompt to compare answers based on
their cited evidence. It returns Answer 1, Answer 2,
or a tie through a structured tool interface.

\begin{lstlisting}[
style=prompt,
caption={System prompt for reference-free answer selection.},
label={lst:selector_system}
]
You are an impartial evaluator selecting the better of two answers to a historical-archives question. You do NOT have a gold answer.

Judge ONLY by which answer is better supported by the Source Passages provided, and by its relevance and completeness for the question.

Rules:
- Base every judgment on the Source Passages. Do NOT use outside knowledge.
- A claim not supported by any passage is a weakness, even if it sounds plausible.
- Do NOT prefer an answer for being longer, more fluent, or more confident.
- If the answers are equally well-supported or indistinguishable, choose "tie".
- Think about grounding first, then completeness, before deciding.
\end{lstlisting}


\end{document}